\documentclass[10pt,conference]{IEEEtran}
\IEEEoverridecommandlockouts
\usepackage{enumitem}
\usepackage{amsmath,amssymb,amsfonts}
\usepackage{algorithmic}
\usepackage{graphicx}
\usepackage{textcomp}
\usepackage{xcolor}
\usepackage[switch]{lineno}

\def\BibTeX{{\rm B\kern-.05em{\sc i\kern-.025em b}\kern-.08em
    T\kern-.1667em\lower.7ex\hbox{E}\kern-.125emX}}

\newcommand{\Comment}[1]{}
\newtheorem{thm}{Theorem}

\newtheorem{require}[thm]{Safety Req.}

\begin{document}
%\linenumbers

\title{Towards Rigorous Design of OoD Detectors\vspace{-0.2em}}

\author{\IEEEauthorblockN{Chih-Hong Cheng}
\IEEEauthorblockA{\textit{Technical University of Munich}\\
Garching, Germany }

\and
\IEEEauthorblockN{Changshun Wu}
\IEEEauthorblockA{\textit{University of Grenoble Alps}\\
Grenoble, France }

\and
\IEEEauthorblockN{Harald Ruess}
\IEEEauthorblockA{\textit{fortiss GmbH}\\
Munich, Germany}
\and
\IEEEauthorblockN{Saddek Bensalem}
\IEEEauthorblockA{\textit{University of Grenoble Alps}\\
Grenoble, France }
}

\maketitle

\begin{abstract}

Out-of-distribution (OoD) detection techniques are
instrumental for safety-related neural networks.
We are arguing, however, that current performance-oriented OoD detection techniques geared towards matching metrics such as expected calibration error, are not sufficient for establishing safety claims.
What is missing is a rigorous design approach for developing, verifying, and validating OoD detectors. 
These design principles need to be aligned 
with the intended functionality and the operational domain. 
Here, we formulate some of the key technical challenges, together with a possible way forward, for developing a rigorous and safety-related design methodology for OoD detectors.

\end{abstract}

%%%%%%%%%%%%%%%%%%%%%%%%%%%%%%%%%%%%%%%%%%%%%%%%%%%%%%%%%%%%%%%%%
%%%%%%%%%%%%%%%%%%%%%%%%%%%%%%%%%%%%%%%%%%%%%%%%%%%%%%%%%%%%%%%%%

\section{Motivation and Reflection on the SotA}
%%%%%%%%%%%%%%%%%%%%%%%%%%%%%%%%%%%%%%%%%%%%%%%%%%%%%%%%%%%%%%%%%
%%%%%%%%%%%%%%%%%%%%%%%%%%%%%%%%%%%%%%%%%%%%%%%%%%%%%%%%%%%%%%%%%

Deep neural networks (DNN) are widely used for vision, control, and natural language processing tasks. For them to be used in \emph{open} and \emph{complex} autonomous systems such as automated driving, one fundamental challenge is to ensure that the intended functionality of the DNN-enabled system does not lead to unacceptable risks. Towards such a challenge, it is widely perceived in the industry that techniques for detecting out-of-distribution (OoD) are the cure, where OoD refers to data points or examples that differ significantly from the training data a model has been exposed to during its training phase. When the model is presented with a data point that significantly deviates from the training distribution, it may struggle to make accurate and reliable predictions.

Although the research on improving OoD detection has advanced significantly these years, one fundamental question remains to be whether the modeling, design, and analysis of OoD detectors are done in a rigorous manner to make them suitable to be used in safety-critical applications. After all, the ambiguous definition of ``\emph{differ significantly from the training data}'' generates concerns such as the appropriateness of evaluation metrics, leaving ample space for software engineering research to be explored. Many of the state-of-the-art OoD detectors are also machine-learned components by design. As they act as the ``checker" in the classical doer-checker safety engineering paradigm, even more engineering rigor is required. In the following sections, we consider how to characterize the requirements for OoD detectors to be used in safety-critical applications and propose methods  in empirical software engineering to be introduced to increase the overall rigor of assurance claims. 

\section{From High-level Safety Requirements to Requirements on OoD Detectors}

\begin{figure*}[t]
    \centering
    \includegraphics[width=0.75\textwidth]{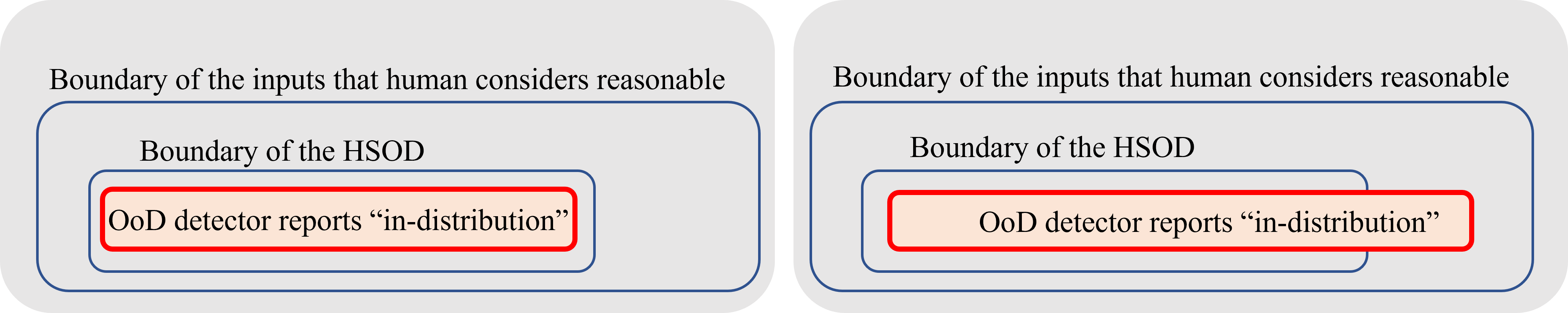}
          \vspace{-3mm}
    \caption{Decision boundary on the OoD detector against the HSOD being appropriate (left) and inappropriate (right) }
    \label{fig:ood.boundary.vs.human}
      \vspace{-3mm}
\end{figure*}

Ensuring the safety of the intended functionality of systems
with learning-enabled DNN components
commonly reduces to demonstrating that
certain error rates are sufficiently low, say, below 
a predetermined threshold~$\epsilon_{err}$,
when operating in a \emph{human-specified operational domain (HSOD)}\@. 
Consider the diagram in Figure~\ref{fig:ood.boundary.vs.human}, where for the complete input space (colored in gray), we may conceptually draw the boundary on a subspace of the input space, where the decision boundary matches the semantics of what human consider reasonable. An example falling outside the boundary can be an image with random noise patterns; in autonomous driving, such an image is not possible to be taken from a camera that is free from hardware and software faults. Within that decision boundary, there is a sub-space as the HSOD, which characterizes the input space that the DNN-enabled system is expected to function. For example, suppose the system is not expected to be operated in snowy weather. In that case, the space formed by the decision boundary on the HSOD shall not include images of snowy weather. This leads to the following safety requirement on the OoD detector to ensure that \emph{it can filter out situations outside HSOD}. 

\vspace{1mm}
\begin{require}\label{req.ood.1}%[REQ.OoD.1]
The set of states where the OoD detector considers ``in-distribution'' should be a subset of the set of states characterized by the HSOD. In other words, the decision boundary of an OoD detector should be at most as large as the decision boundary as characterized by the HSOD, as illustrated in the left side of Figure~\ref{fig:ood.boundary.vs.human}. 
\end{require}

\begin{figure*}[t]
    \centering
    \includegraphics[width=0.75\textwidth]{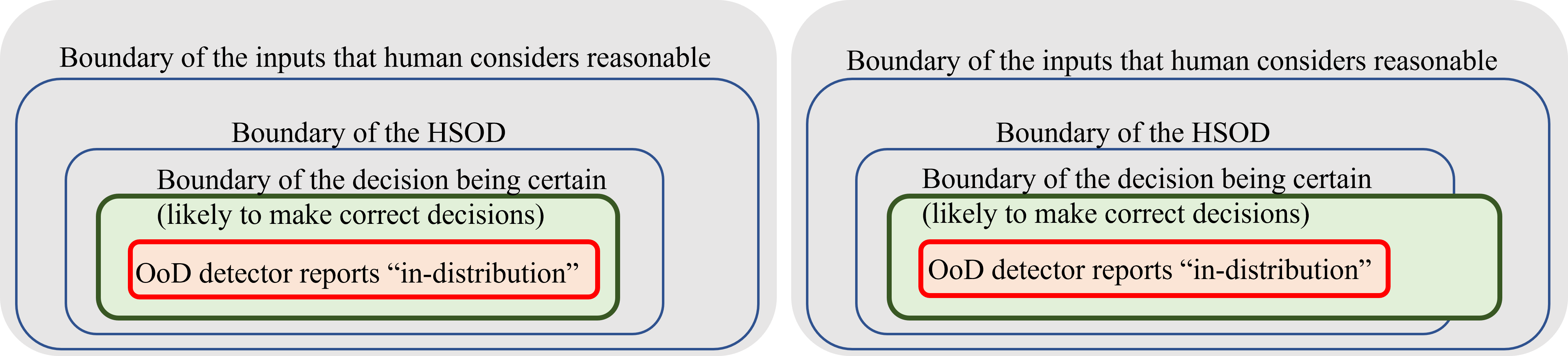}
          \vspace{-3mm}
    \caption{Decision boundary on the OoD detector against the conceptual decision boundary where the DNN is certain }
    \label{fig:ood.boundary.vs.correct}
    \vspace{-3mm}
\end{figure*}

\vspace{1mm}
Subsequently, let us consider the following (more realistic) safety goals reflected as requirements. 

\vspace{1mm}
\begin{require}\label{req.uncertain}
If the DNN-based system is uncertain about its prediction, and it reports ``uncertain'', then the output of the system is considered to be correct. 
\end{require}

\begin{require}\label{req.certain}
If the DNN-based system is certain about its prediction, then it should generate correct predictions with an error rate less than the pre-defined threshold~$\epsilon_{err}$. 
\end{require}

\vspace{1mm}

With Requirement~\ref{req.uncertain} and~\ref{req.certain}, one can infer that the safety requirement on the error rate less than the threshold is satisfied (due to case split), but again this is not a requirement on the OoD detector; the DNN can be uncertain in its prediction, but the input data is “close” to the sampled distribution from the view of the OoD detector. Thus we are suggesting the following requirement on restricting the decision boundary of OoD (i.e., the definition of being ``close'') to be inside the decision boundary of ``DNN being certain''.

\vspace{1mm}

\begin{require}\label{req.ood.2}
If the DNN is uncertain about its prediction, then the OoD detector shall report that the input is ``not-in-distribution''. 
That is, the set of states where the OoD detector considers ``in-distribution'' should be a subset of the set of states where the DNN is certain, as illustrated in Figure~\ref{fig:ood.boundary.vs.correct}.

\end{require}

\vspace{1mm}

As one can observe from Figure~\ref{fig:ood.boundary.vs.correct}, the introduced Requirement~\ref{req.ood.1} and~\ref{req.ood.2} do not enforce a relation between the decision boundary between the HSOD and the conceptual decision boundary where the DNN is certain about the prediction. This altogether implies that for the OoD detector to be used in safety-critical applications, these two requirements, characterized by set containment relations, can be checked independently. This also means that OoD methods do not necessarily need to produce uncertainty quantification. 

Consider the case where we set the decision threshold of the OoD detector for reporting ``out-of-distribution'' to be~$\kappa$. It may suffice to provide evidence that within the decision boundary of $\kappa$ (i.e., ``in-distribution''), the DNN has an error rate that is less than~$\epsilon_{err}$. Contrarily, satisfaction on the expected uncertainty error (ECE) metric can not provide a direct safety assurance claim and can easily lead to unnecessary over-engineering\footnote{For example, for DNN predicting the results with $\alpha\%$ confidence, then the satisfaction of the ECE metrics aims to show that, on average, $100-\alpha$ out of~$100$ prediction mistakes are expected or observed in the experiments. In practice one separates the into histograms with each bin characterizing the result for predictions with $0-10\%, 10-20\% \ldots, 90-100\%$ confidence. }. 

\vspace{-1mm}
\section{Practical Considerations}

Formally proving such a requirement can be difficult in the field of machine learning. Using statistical methods such as hypothesis testing (as successfully demonstrated in the field of empirical software engineering and the medical domain) can serve as a method to derive confidence measures by associating with the number of populations being used in testing. By combining the confidence generated in each piece of evidence with the logical assurance claim, altogether it enables a solid understanding regarding the effectiveness of OoD detectors, following the rigorous confidence propagation methods utilizing  Dempster-Shafer theory~\cite{shafer1976mathematical,wang2019safety,cheng2022logically}. We conclude by enumerating three challenges to realize the concept.

\textbf{(C1)} To perform hypothesis testing, the first challenge is establishing the sample from which a statistical analysis is created for a study. For image-based object detection, under the premise that the image formation is free from errors, one may use all data being collected in an extended duration with the same configuration (e.g., fleet operated for a year) as the population. The data collection process should ensure that natural data outside the HSOD is also captured to perform bug finding. As demonstrated in Figure~\ref{fig:ood.boundary.vs.human}, an OoD detector makes a mistake, if it considers a data point outside HSOD to be in-distribution. For autonomous driving, such a process-related requirement is actionable when the sensory system is first deployed on vehicles as a driver assistance feature (where the vehicle owner may use the vehicle in all reasonable situations), followed by ``feature upgrading'' to autonomous driving, where the HSOD is made explicitly restricted.

\textbf{(C2)} Another fundamental challenge is the decision boundary regarding ``\emph{DNN generating a prediction being certain}". Practically, one does not need to know the exact boundary, as one only uses sampled data points to test the hypothesis of the set inclusion relation being satisfied. The question arises: How can one objectively\footnote{By ``Objective'', what we mean is that the decision is not based on the (un)certainty information provided by the model, as the model can sometimes be overly confident.} know the (un)certainty of a DNN predicting an input data point~$\vec{p}$? To mediate this issue, apart from correctness on the particular input data point~$\vec{p}$, one may define the point-wise certainty by using the DNN to predict a set of nearby data points $\{\vec{p} + \vec{\delta} \;|\; \lVert\vec{\delta}\rVert \leq \Delta\}$, with $\vec{\delta}$ referring to  a small perturbation (generated randomly with quantity less than~$\Delta$) on~$\vec{p}$. This idea of objectively quantifying the certainty connects to the randomized smoothing technique used in defending adversarial examples~\cite{cohen2019certified}.

\textbf{(C3)} Finally, with Requirement~\ref{req.ood.1} and~\ref{req.ood.2} as guiding principles, one should also consider how DNNs, OoD detectors and the HSOD should be specified and designed to increase the likelihood of fulfilling the requirements. As an example, if the HSOD is bound to a geographical location such as Munich, standard ML-based OoD detectors are unlikely to differentiate if an image is taken in Munich or other cities in Germany. The specification of the HSOD should thus focus on characterizing the semantic attributes (e.g., features of pedestrians or weather conditions) that enable analyzability.

\bibliographystyle{IEEEtran}
%\bibliography{ref}


\end{document}